\documentclass[tablecaption=bottom]{jmlr}%

\usepackage{graphicx}
\usepackage{sidecap}
\usepackage{xspace}

\usepackage{color}
\usepackage{xcolor}
\usepackage{hyperref}
\usepackage{tablefootnote}
\usepackage{enumitem}
\usepackage{longtable}

\usepackage{booktabs}
\usepackage[load-configurations=version-1]{siunitx} %

\theorembodyfont{\upshape}
\theoremheaderfont{\scshape}
\theorempostheader{:}
\theoremsep{\newline}

\jmlrvolume{1}
\jmlryear{2021}
\jmlrsubmitted{submission date}
\jmlrpublished{publication date}
\jmlrworkshop{NeurIPS 2020 Competition Track} %

\title{NeurIPS 2020 NLC2CMD Competition:\\ Translating Natural Language to Bash Commands}

\author{\Name{All Contributors}}

\author{
\Name{Mayank Agarwal} \Email{mayank.agarwal@ibm.com}\\
\addr IBM Research
\AND
\Name{Tathagata Chakraborti} \Email{tchakra2@ibm.com}\\
\addr IBM Research
\AND
\Name{Quchen Fu} \Email{quchen.fu@vanderbilt.edu}\\
\addr Vanderbilt University
\AND
\Name{David Gros} \Email{dgros@ucdavis.edu}\\
\addr University of California, Davis
\AND
\Name{Xi Victoria Lin}\thanks{Work done while the author was at Salesforce Research.} \Email{victorialin@fb.com}\\
\addr Facebook AI
\AND
\Name{Jaron Maene} \Email{jaron.maene@gmail.com}\\
\addr KU Leuven
\AND
\Name{Kartik Talamadupula} \Email{krtalamad@us.ibm.com}\\
\addr IBM Research
\AND
\Name{Zhongwei Teng} \Email{zhongwei.teng@vanderbilt.edu}\\
\addr Vanderbilt University
\AND
\Name{Jules White} \Email{jules.white@vanderbilt.edu}\\
\addr Vanderbilt University
}

\newcommand{\nlctocmd}{\texttt{NLC2CMD\xspace}}
\newcommand{\bash}{Bash\xspace}

\newcommand{\nlc}{{\tt NLC2CMD}}

\begin{document}

\maketitle

\vspace{-2em}
\begin{abstract}
The \nlctocmd~Competition hosted at NeurIPS 2020 aimed to bring 
the power of natural language processing to the command line. 
Participants were tasked with building models that can transform 
descriptions of command line tasks in English to their Bash syntax.
This is a report on the competition with details of the task, metrics,
data, attempted solutions, and lessons learned.
\end{abstract}

\vspace{5pt}
\begin{keywords}
Natural Language Processing, Bash, Programming, Code, Competition 
\end{keywords}

\section{Introduction}
\label{sec:intro}

The command line interface (CLI) has long been 
an invaluable computing tool
due to its %
expressiveness, efficiency, and extensibility.
While graphical user interfaces (GUIs) %
struggle to keep up with the %
fast changing features in software development
(for example, consider the time it took to move from Docker to
Kubernetes %
in cloud platforms), CLIs provide a universal interface to almost any software feature via a combination of text-based commands and command arguments.
Natural language based CLI interaction, if successful over a variety of command line tasks,
has the potential to transform the way we interact with different operating systems and cloud
platforms, %
lowering the barriers to entry 
and increasing the accessibility
of compute resources across the planet. 

The \nlc\ competition revolved around the simple use case:
translating natural language (NL) descriptions of command line tasks 
to their %
corresponding command line syntax, thereby making the CLIs more accessible and more intuitive to use\footnote{A former position paper by the competition organizers on bringing the power of AI to the CLIs can be found in~\cite{agarwal2020clai}.}.
We %
compiled a large dataset consisting of the English (natural language) descriptions
of tasks %
the users want to perform and their corresponding
Bash\footnote{\url{https://www.gnu.org/software/bash/}} invocations (\S~\ref{section:data_sources}). 
The participants needed to build models and systems (\S~\ref{sec:competing-solutions}) that generated the correct Bash command given new and unseen natural language invocations.
To better measure the progress of this task and facilitate cross comparison of the submitted systems, we proposed new accuracy and energy efficiency metrics (\S~\ref{sec:metrics}). We also conducted a careful analysis of the winning systems and shared our suggestions for future work (\S~\ref{sec:discussion}).

\begin{figure}[tbp!]
\centering
\includegraphics[width=0.8\textwidth]{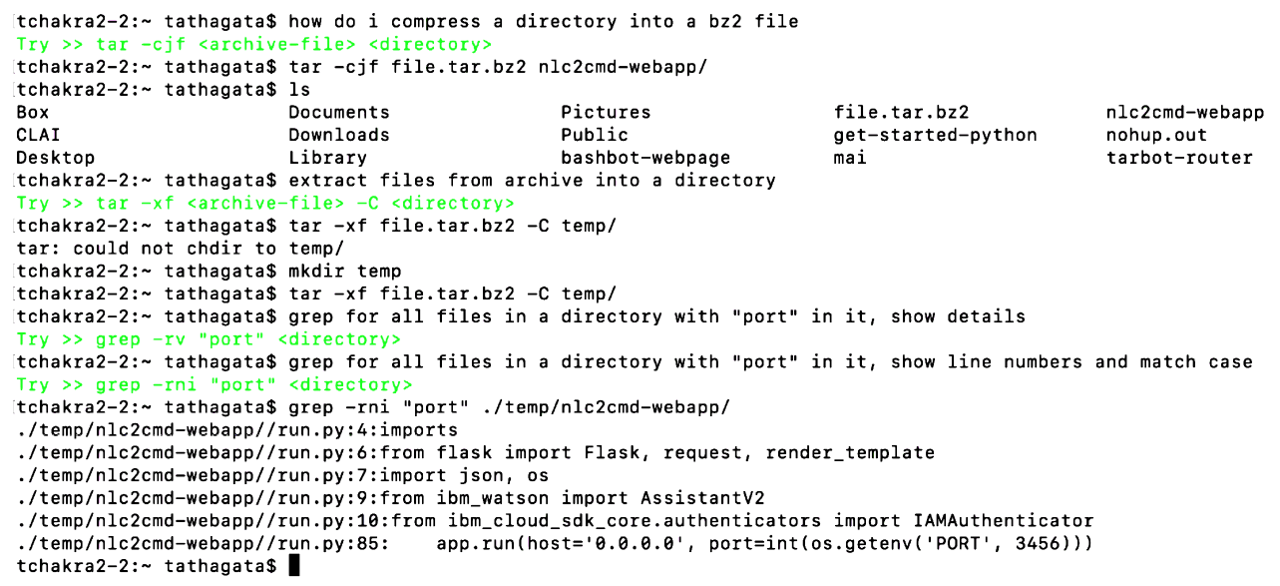}
\caption{An example of %
a command line interface with natural language support,
as described in \cite{agarwal2020clai}.
Note that for normal Bash commands, execution proceeds as usual, while for
tasks described in natural language, an \nlc\ plugin intervenes
(shown in \textcolor{green}{green}) with  
the translation of the described task into the corresponding command 
line syntax.}
\label{fig:nlc2cmd4clai}
\end{figure}

\section{Task Description}

The \nlctocmd\ task aims to translate the natural language description ($nlc$) of a command line task to its corresponding \bash command ($c$). The algorithm ($A$) is expected to model the top-k \bash translations given the natural language description.

$$A~:~nlc \mapsto \{~p~|~p = \langle c, \delta \rangle~\}; ~~~~ |A(nlc)| \leq k$$

Each prediction from the model is a set of \bash commands along with the %
confidence $\delta$ ($0.0 \leq \delta \leq 1.0$) in its prediction. 
In practice, this confidence estimation can be used to filter out uncertain predictions, and it is factored into %
the competition evaluation. By default, the confidence score is set to 1.0, %
indicating full confidence in a model's prediction.

\section{\nlctocmd: Competition Overview}

The competition began in July 2020 and concluded at the end of November 2020. It was divided into three phases: Training, Validation, and Test. %
The Validation phase ran in October and the Test Phase ran in November. 
A total of $20$ teams from around the world signed up for the competition by sending in a signed {\tt Terms \& Conditions} document - a pre-requisite for entry. Among these, $13$ teams remained active through the end of the first phase. 
9 teams went through the end of the Test Phase, and 6 of them open-sourced their solutions.
The teams were allowed a maximum of $100$ submissions in the first two phases, and a maximum of $10$ submissions for the final phase; there were also limits on the number of daily submissions ($5$ for the first 2 phases, and $1$ for the final phase). We hosted the competition using the EvalAI platform~\citep{EvalAI}.\footnote{\url{https://evalai.cloudcv.org/web/challenges/challenge-page/674/overview}}

\section{Data}
\label{section:data_sources}

\subsection{NL2Bash}

The NL2Bash dataset is a supervised natural language to Bash command dataset collected by \cite{LinWZE2018:NL2Bash}. It contains approximately 10k pairs of natural language task invocations and their corresponding command line syntax, covering over 100 commonly used \bash utilities. 

\subsection{Tellina Query Log}

Tellina \citep{LinWPVZE2017:TR} is a recurrent neural network based architecture to translate natural language utterances to \bash commands. The system is also available as a web-based application to the public at large, and has been collecting data of natural language utterances from its users for quite some time now\footnote{\url{http://tellina.rocks/}}. We collected nearly $1000$ natural language utterances recorded from users' interactions with the Tellina system. Three programmers with prior \bash experience then annotated these collected examples: either through their own background knowledge, or by referring to the manual pages, or by searching over the Internet for guidance. Through this process, we generated multiple ground truth labels for a majority of the examples in this dataset; this was beneficial for the evaluation process since the metric used in the competition (Section~\ref{sec:nlc2cmd:metric}) uses the maximum score over the predicted and the ground truth commands.
Once this dataset was annotated, we filtered the data samples through the data filtering process described in section \ref{sec:data-partitions}.
After filtering, we were able to add around $700$ examples to the \nlctocmd\ evaluation dataset.

\subsection{\nlctocmd~Data Collection Track}

In addition to the main challenge of building an algorithm to translate a natural language utterance to a \bash command, we also ran a parallel data-collection track to collect natural language to bash command pairs. These pairs were supplemental to the previous $2$ data sources, and were collected through a web interface hosted on the competition website. Participants in this challenge were asked to submit as many pairs of English invocations and \bash commands as they wished. $21$ participants from industry and academia submitted over $120$ examples in this challenge. These pairs were filtered using the same process as used for the Tellina dataset, and the resulting examples (nearly $100$) were part of the final phase of the challenge.

\subsection{Data partitions and pipeline}
\label{sec:data-partitions}

\paragraph{Data filtering} We validate and filter the data collected through NL2Bash, Tellina Query Log, and the \nlctocmd\ Data Collection Track using the \bash parser used throughout the competition\footnote{\url{https://github.com/IBM/clai/tree/nlc2cmd/utils/bashlint}}. For each data sample pair of natural language invocation and its corresponding \bash command text, we parse the \bash command text through the parser to get its abstract syntax tree (AST) and then reconstruct the command text from the AST. We validate that the reconstructed command matches exactly to the original command text. This process ensures that any text which is not a valid \bash command -- by either specifying a utility which is not part of the parser vocabulary, incorrect flags for the corresponding utility, or incorrect syntax altogether -- is identified and omitted from the dataset.
We also ensure that for a given command's text, all the utilities included in it are part of the Ubuntu 18.04 LTS command set\footnote{\url{http://manpages.ubuntu.com/manpages/bionic/en/}}. Any command text that includes any utility that is not in the Ubuntu utility set is omitted.

\paragraph{Training} For training, participants were provided with a filtered version of the \texttt{NL2Bash} \citep{LinWZE2018:NL2Bash} dataset, which is a supervised dataset of natural language descriptions mapped to their respective \bash commands. Participants were also provided with %
a version of the manual (man) pages for Ubuntu 18.04 LTS\footnote{\url{http://manpages.ubuntu.com/manpages/bionic/en/}} to enable models to utilize information available in these %
man pages to learn and suggest commands unseen in the structured dataset. In addition to these two datasets, participants were free to use any freely-available dataset that they deemed fit to train their models, conditioned upon its disclosure and release to all competitors and the general public before the final stage of the competition.

\begin{figure}[tbp!]                                                       \subfigure[Test dataset sizes across the 3 phases]{  
\includegraphics[width=0.26\textwidth]{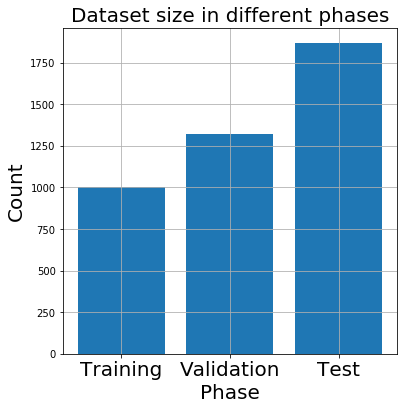}     \label{fig:dataset_sizes}
} 
\hfill
\subfigure[Most frequent utilities: validation phase dataset]{ 
\includegraphics[width=0.33\textwidth]{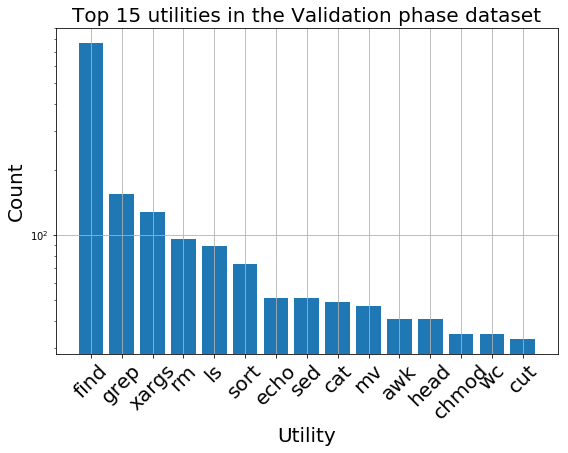}          \label{fig:val-phase-dist}
}
\hfill
\subfigure[Most frequent utilities: test phase dataset]{
\includegraphics[width=0.33\textwidth]{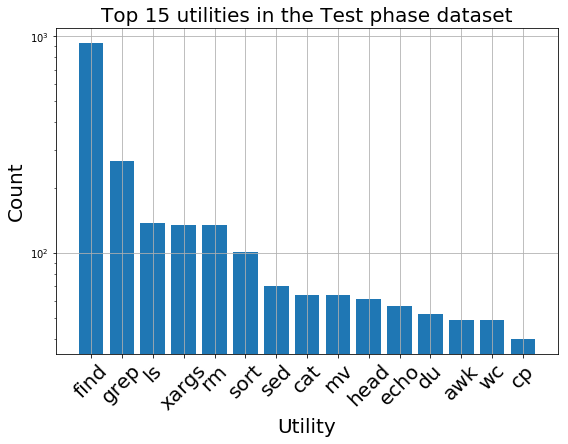}          \label{fig:test-phase-dist}                                                 }                                                      
\caption{                                                                      
Dataset characteristics in the \nlctocmd\ competition
}                                                                              
\label{fig:path}                                                               
\end{figure}

\paragraph{Validation and Test} We used a combination of data collected from sources described 
previously in this section 
to create the evaluation dataset. 
We create the test set for the first phase of the competition by randomly sampling $1000$ data samples from a filtered version
of the complete NL2Bash dataset \citep{LinWZE2018:NL2Bash} (explained in detail in the {\em Data filtering} paragraph above). This data split %
is different from that of the original NL2Bash data release; however, there may be some overlap in the data samples. The primary difference between the two datasets lies in the different choices of random seed and the number of test examples sampled.

In the subsequent phases, we add new data samples to the previous phase test set to create the test set of the following phase.
$324$ new data samples from the Tellina query log were added to create the the \texttt{validation} phase test set. To create the test set of the final phase of the \nlc\ competition, $543$ additional data samples from the Tellina query log and the \nlc\ challenge were added to the test set of the \texttt{validation} phase.
Through the different phases of the competition, the size of the evaluation dataset grew significantly. With the addition of data from the Tellina query log and the \nlctocmd~challenge, the evaluation data size increased from $1000$ samples in the \texttt{training} phase to over $1800$ in the \texttt{test} phase (see figure \ref{fig:dataset_sizes}).

To validate for algorithms' performances on Bash utilities not seen in the training data, we also added utilities which were not part of the NL2Bash training data in the \texttt{test} phase of the competition. While there were $101$ utilities in the first phase of the competition, $27$ new utilities were added through the Tellina data and the \nlctocmd~challenge to bring the total to $128$ utilities in the final phase. A distribution of the count of the top $15$ utilities in the validation and test phase data samples is displayed in Figures \ref{fig:val-phase-dist} and \ref{fig:test-phase-dist} respectively.

We note that due to: (i) the random subsampling of the NL2Bash dataset to create the test set for the first phase of the competition; and (ii) the addition of data samples to this existing set to create the test set for subsequent phases -- there is a possibility of algorithms overfitting by training on the publicly available NL2Bash dataset and inflating their scores. However, we decided on this format for the following reasons: a) NL2Bash is the only dataset available for the specific \nlc\ task; since the other two data sources (Tellina query log, and \nlc\ data challenge) weren't available right at the beginning of the competition, we had to utilize NL2Bash to create the initial test set in order to adhere to the competition schedule; and b) Adding samples to the initial test set to create the test set for the subsequent phases (instead of switching completely to the newly collected data samples) allowed us to control and prevent the underlying data distribution from shifting significantly between competition phases.

\section{Metrics}
\label{sec:metrics}

We evaluated submissions to the \nlctocmd\ competition on two metrics: Accuracy, and Energy Consumption. In the following, we delve deeper into the specifics and the philosophy behind using two metrics to evaluate submissions to the competition.

\subsection{Accuracy}
\label{section:acc}

In this section, we first discuss existing 
metrics used in prior work
and their shortcomings in evaluating a structured prediction task such as \nlctocmd. We subsequently discuss a metric that verifies by execution. This metric fixes many shortcomings of the existing accuracy metrics, but requires considerable effort to implement. We briefly discuss the requirements to make this evaluation metric practical. Finally, we describe the metric we proposed for 
the competition.

\subsubsection{Existing metrics}
\label{sec:existing-metrics}

\paragraph{Full Command Accuracy:} Accuracy metric or measuring exact match between the generated and reference sample has precedence in code generation research \citep{yin2017syntactic, chen2018tree, fu2019coda}. Specifically, the Full Command Accuracy metric was utilized by \cite{LinWPVZE2017:TR, LinWZE2018:NL2Bash} in their work on synthesizing bash commands given their natural language descriptions. The full command accuracy $\text{Acc}_{F}^{k}$ is defined as the percentage of test instances for which a correct full command is ranked $k$ or above in the model output. 
Because this metric measures the correctness of the complete command -- command utility, utility flags, and the command ordering -- it has a stricter notion of correctness. However, since \bash is a turing-complete language \citep{wikibooks:bash}, verifying the equivalence of two commands is undecidable \citep{LinWZE2018:NL2Bash}; and therefore the chance of accurately measuring the correctness of the predicted command depends on how exhaustive the test set is. Theoretically, if the test set contains all the possible ways of achieving the task, this metric would accurately measure the predictors performance. However, this can be difficult, which is why the work of \cite{LinWZE2018:NL2Bash} relied on human evaluations along with using this metric to measure their system performance.

\paragraph{BLEU score:} BLEU score \citep{papineni2002bleu} is a widely-used automatic machine translation evaluation metric that serves as an alternative to the otherwise time-consuming and expensive human evaluation. BLEU score is computed by comparing the n-grams of the candidate translations with the n-grams of the reference translation. 
We refer the reader to the work of \cite{gatt2018survey} for a detailed survey of evaluation methodologies in natural language generation and their characteristics.

\cite{tran2019does} study the effectiveness of using the BLEU score to evaluate the task of code migration, and conclude that the metric is not effective in reflecting the semantic accuracy of translated code. Additionally, they found a weak correlation between the BLEU score and human judgement of the semantic correctness of the translated code. Similar results were observed by \cite{LinWPVZE2017:TR} in their study of using the BLEU score for the task of converting natural language to its corresponding bash syntax, and by \cite{allamanis2018survey} in their broader work on the intersecting research areas of Machine Learning, Programming Languages, and Software Engineering.

\paragraph{Template Accuracy:} The Full Command Accuracy matches the complete predicted command with the corresponding reference commands to measure the performance of the predictor. The full command includes the individual utilities that compose the command, along with their flags, arguments, and the ordering of the utilities in case the command is a composition of multiple utilities. While ideally it is desirable to predict the complete command, in the normal development workflow, predicting the correct utilities and their flags is of greater importance than predicting the arguments as well. This is evidenced by the current user workflow, where the user searches manual pages, online forums, etc. for a way to accomplish their task, and then fills in their desired arguments once they've identified the appropriate utilities. Specifying the entire arguments is sometimes inconvenient for the user as well -- consider specifying the full paths of files in case the user wants to move files from one location to another, when asking for only the utility is much easier in this case. To accommodate for this relative importance, \cite{LinWZE2018:NL2Bash} propose the Template accuracy metric $\text{Acc}_{T}^{k}$. 
This is defined to be the percentage of test instances for which a correct command template is ranked $k$ or above in the model output (i.e., ignoring incorrect arguments), and is also calculated via human evaluation \citep{LinWZE2018:NL2Bash}.

\subsubsection{Verification by Execution}
\label{sec:verification-by-execution}

Since \bash is a turing-complete language, verifying the equivalence of two commands to achieve the same task is undecidable~\citep{10.1145/3314221.3314596}. Consider a simple case of listing all the files under a folder recursively. The 3 commands \texttt{tree /dirpath/}; \texttt{ls -R /dirpath/}; and \texttt{find /dirpath -print} all achieve the desired purpose but use different utilities. %
With the same utility, the commands could still differ by using different sets of flags.
To account for such variation, we could execute the predicted command and the reference in a controlled shell environment and match the pre and post-execution outputs of the two commands\footnote{This approach tends to overestimate accuracy, as a command and the reference may execute to the same results in a particular test environment while being semantically different. For example, if a directory contains only \texttt{.txt} files, \texttt{rm .} and \texttt{rm . -name ``*.txt''} will both leave an empty folder after execution.~\cite{DBLP:journals/corr/abs-2010-02840} mitigates this issue by testing the programs in multiple environments, thereby reducing the chance of a false positive.}. A similar evaluation setup (verification by execution) is utilized by \cite{lachaux2020unsupervised} in their work on translation between programming languages; by \cite{zhong2017seq2sql} and \cite{DBLP:journals/corr/abs-2010-02840}  when mapping natural language to SQL queries; and by \cite{guu2017language} in their work on mapping natural language to logical forms in a synthetic setting.

The verification by execution setup %
requires a sand-boxed \bash environment where commands can run in a reproducible manner and without interference from other processes running on the machine. 
The test data is preferably command-specific to effectively rule out false positives.
In addition, it is challenging to measure and compare the side effects\footnote{\url{https://en.wikipedia.org/wiki/Side_effect_(computer_science)}} of the commands.

\subsubsection{\nlctocmd\ metric}
\label{sec:nlc2cmd:metric}

Having introduced existing accuracy metrics and their drawbacks in Section \ref{sec:existing-metrics}, we propose a metric that: a) ignores arguments in the predicted commands; b) considers the order of the utilities in case of piped commands; and c) penalizes predicting utility flags in excess of those defined in the ground truth commands. 

We define $U(c)_i$ to be the $i^{th}$ utility in a command (\texttt{c}), and $F(u)$ to be a set of flags for a utility $u$. Given the predicted command $C_{\text{pred}}$ and the reference command $C_{\text{ref}}$, we first define the flag score $S_{F}^{i}$ to be:

\begin{equation}
S_{F}^{i} (C_{\text{pred}}, C_{\text{ref}}) = \frac{1}{N} \Big( 2 \times |F(U(C_{\text{pred}})_i) \cap F(U(C_{\text{ref}})_i)| - |F(U(C_{\text{pred}})_i) \cup F(U(C_{\text{ref}})_i)| \Big)
\end{equation}

Here, $N = \max \big( |F(U(C_{\text{pred}})_i))|, |F(U(C_{\text{ref}})_i)| \big)$. The flag score rewards flags common to both the predicted and the reference command, but also penalizes  excess predicted flags.

The normalized score for a single prediction $p = \langle C_{\text{pred}}, \delta \rangle$ is then computed as:

\begin{equation}
S(p) =  \max_{C_{\text{ref}}} \sum_{i \in [1, T]} \frac{\delta}{T} \times \bigg( \mathbb{I}[ U(C_{\text{pred}})_i = U(C_{\text{ref}})_i ] \times \frac{1}{2} \Big(1 + S_{F}^{i} \Big) - \mathbb{I}[ U(C_{\text{pred}})_i \not= U(C_{\text{ref}})_i ] \bigg)
\end{equation}

Here, $T = \max \big( |U(C_{\text{pred}})|, |U(C_{\text{ref}})| \big)$, and $\mathbb{I[\cdot]}$ is the indicator function.

Since the algorithm $A(nlc)$ predicts more than one candidate command to the natural language query $nlc$, the overall score is computed as:

\begin{equation}
Score(A(nlc))= 
\begin{cases}
\max_{p \in A(nlc)} S(p),   & \text{if } \exists_{p \in A(nlc)} \text{ such that } S(p) > 0\\
\frac{1}{|A(nlc)|} \sum_{p \in A(nlc)} S(p),              & \text{otherwise}
\end{cases} 
\end{equation}

This scoring mechanism incentivizes the precision and recall of the correct utility and its flags, weighted by the algorithm's reported confidence.\footnote{
Note that an initial version of the metric
naively computed the maximum score for the top-$k$ predictions 
from the algorithm. Since the metric considers both the predicted command 
as well as the confidence measure, it was possible to artificially 
clamp the metric to zero instead of its normal range of $[-1, 1]$. This could be
achieved by predicting the last of the $k$ commands with $0.0$ confidence, thus
ensuring a score of $0.0$ for the final prediction; and clamping the overall
metric minimum to $0.0$ as well. This was fixed by modifying the aggregation of the 
individual scores of the metric by using a \texttt{max} operator only if
at least one of the prediction scores is greater than $0$; and an average otherwise.
}
The reasons for choosing this metric are many:
1) It can be evaluated within the
the timeline of a competition at a conference;\footnote{
While similar competitions have looked to implement unit tests 
for the test data before \citep{hendrycks2021measuring},
for us, the test set was not fixed apriori but only determined during the course of the NLC2CMD Challenge; this, along with the fact that the testing scheme for the competition without arguments leads to commands that are under-determined, meant that the simulation option was out of scope.} 
2) Synthesizing the full command is a more difficult task and we wanted to start from a simpler option given this is the first time the competition is running; 
3) In some cases it is challenging to fully specify the command arguments in the natural language;
4) A system that suggests a command with argument placeholders can be useful since once the command structure is known, the user may fill the missing argument values in an interactive manner with an AI assistant on the terminal like CLAI~\citep{agarwal2020clai}. We provide examples illustrating the behavior of this metric under different ground truth and predicted command variations in appendix \ref{sec:appendix:metric:examples}.

\subsection{Energy Efficiency}

Evaluating and mitigating the energy consumption of deep learning models has recently gained traction in the research community. \cite{li2016evaluating} studied the energy efficiency of convolutional neural networks, while more recently, the work of \cite{strubell2019energy} evaluated the energy consumption of developing deep learning models for NLP tasks. \cite{strubell2019energy}'s findings suggested that the $\text{CO}_2$ emissions from training an NLP pipeline (with tuning and experimentation) is greater than the average yearly emission of an American life. In the same spirit, the work of \cite{schwartz2019green} distinguish between Red AI and Green AI. Red AI refers to the pursuit of obtaining state-of-the-art results in accuracy (or related measures) through the use of massive computational power. In contrast, Green AI refers to AI research that yields novel results without increasing computational cost, and ideally reducing it. The authors analyzed 60 randomly sampled papers from the ACL 2018\footnote{\url{https://acl2018.org/}}, CVPR 2019\footnote{\url{https://cvpr2019.thecvf.com/}}, and NeurIPS 2018\footnote{\url{https://nips.cc/Conferences/2018}} conferences to find that a large majority of papers target accuracy or some related measure, and thus, \cite{schwartz2019green} advocate for making efficiency a research criteria along with accuracy to make AI greener. Their measures of efficiency include carbon emissions, electricity usage, among others. Since then, \cite{lottick2019energy}, \cite{lacoste2019quantifying}, and \cite{henderson2020systematic} have proposed tools to help researchers evaluate and report the energy consumption requirements of their work. Additionally, recent initiatives such as the SustainNLP\footnote{\url{https://sites.google.com/view/sustainlp2020/}} workshop at EMNLP 2020\footnote{\url{https://2020.emnlp.org/}} have focused on encouraging development of efficient NLP models, and providing simpler model architectures and empirical justification of model complexity.

While the aforementioned work focuses on the energy consumption of models during the training phase, with the deployment of these models, their energy consumption in inference phase can outweigh their training cost over time. This dichotomy between the energy cost during training in machine learning research and the energy cost during inference in production systems, and the need to also measure the inference energy cost is highlighted by \cite{henderson2020systematic} and \cite{bender2021dangers}. 
In accordance with the recommendations of Green AI and other recent work in the energy-efficient machine learning research, we measure the energy consumption of submissions to the \nlctocmd\ challenge using the \texttt{experiment-impact-tracker}\footnote{\url{https://github.com/Breakend/experiment-impact-tracker}} library by \cite{henderson2020systematic}. The Efficiency track at the \nlctocmd\ competition was run in parallel to the Accuracy track and aimed to award the most energy-efficient model amongst the top-5 scoring (by accuracy) models.

\section{Competing Solutions}
\label{sec:competing-solutions}

Table~\ref{tab:leaderboard} provides a snapshot of the \nlc\ competition leaderboard at the end of the final (test) phase: a total of $6$ teams/entries and $2$ baselines made it to this stage.\footnote{Note that the Tellina was used ``as is'' for the competition as a baseline and was not retrained for the new metric so as to not make the competition task (running for the first time) too hard for participants.} The leaderboard captured the accuracy score, energy consumption, and latency of the entries. We also provide examples of the natural language invocation along with the ground truth command and the commands predicted by the top 3 teams in appendix \ref{sec:appendix:model:outputs}. In the following, we provide brief overviews of the techniques adopted by these entries.

\begin{table}[]
\parbox{.60\linewidth}{
\tiny
\centering
\begin{tabular}{cccc}
\toprule
Team Name & Accuracy Score & Energy (Joules) & Latency (sec) \\ \midrule
\textbf{Magnum} & \textbf{0.532} & 0.484 & 0.709 \\
\textbf{Hubris} & \textbf{0.513} & 12.037 & 14.870 \\
Jb & 0.499 & 2.604 & 3.142 \\
\textbf{AICore} & 0.489 & \textbf{0.252} & 0.423 \\
AINixCLAISimple & 0.429 & N.A. & 0.010 \\
coinse-team & 0.163 & 343.577 & 0.452 \\ \hline
Tellina & 0.138 & 2.969 & 3.242 \\ \hline
\end{tabular}%
\caption{Final leaderboard for the \nlc\ competition, 
showing the accuracy score for the final (test) phase, along with the energy consumed and latency for every invocation.
}
\label{tab:leaderboard}
}
\hfill
\parbox{.37\linewidth}{
\tiny
\centering
\begin{tabular}{l c} 
\toprule
IR-Baseline Variation & Accuracy Score \\  
\midrule
Tf-IDF Raw & 0.361\\
+ AInix Data        & 0.404\\
+ Prune Duplicates  & 0.413\\
+ Normalize NL      & 0.429\\
+ Adjust Conf.      & 0.472\\
\bottomrule
\end{tabular}
\caption{Results from simple IR baselines. Additions to the raw predictor are retained cumulatively top-to-bottom.}
\label{tfidfvariations}
}
\end{table}

\subsection{TF/IDF and Proposed New Baselines}

Team {\tt AINixCLAISimple} briefly explored some simple baselines on the task. Exploration of simple baselines might evaluate if the metrics or task are too easily ``gameable". This team did not find a simple technique that could achieve the highest accuracy score, but did vastly outperform the original host Tellina baseline at very high efficiency.

The primary explored approach was an information retrieval (IR) approach using Tf-IDF rankings. Several variations were explored (\autoref{tfidfvariations}). In all variations the NL-CMD training pairs were first indexed using the Vec4IR package \citep{mci/Galke2017}. In the simplest variation (\texttt{Tf-IDF Raw}) the team took the evaluation utterances, queried the training, and returned the top 5 highest Tf-IDF matches each with a predicted confidence of $1.0$. This approach had a score of 0.361, which is approximately 2.6x higher than the Tellina baseline.

Several additions were made. The inclusion of the AInix Archie data\footnote{\url{https://github.com/DNGros/ai-nix-kernal-dataset-archie-json}} as training data boosts the score to 0.404. Additionally, the team notes there might be duplicates of a given command in the training data (especially considering the constant arguments of the command are ignored). Because the metric takes a max, it is not reasonable to include duplicates in the five outputs. A gain of about 0.01 is achieved by pruning duplicates, instead taking the next ranked retrieved commands, up to five unique commands. 
A further 0.016 points can be gained by applying some string matching heuristics to normalize NL tokens which appear to be constants (file names, numbers, IP addresses, etc). 

Finally, an approach was used to lower points deductions on difficult NL utterances. A logistic regression model was fit to predict the probability the prediction would result in a positive score. It used features like the Tf-IDF score, number of flags in the retrieved command, and number of utilities. These logistic regressions were used in combination with the estimated probability that all predictions will score negatively in order to potentially lower the outputted confidence to be less than $1.0$. This resulted in a 0.043 point gain. 

This team also explored a method of ``optimally diverse picking". This was based off the observation that the metric considered the maximum score, so it might be optimal to return a diverse range of commands in the hopes at least one receives a positive score. Various approaches were explored for optimizing pick diversity, but it was found that when paired the Tf-IDF model with poorly calibrated score estimates, this approach did not help performance. A better underlying model might make the approach more useful.

The best performing model under the Tf-IDF approach achieved an accuracy score of 0.472.\footnote{Note, due to participant user error, the ``+ Adjusted Conf" model failed to upload to the evaluation server; thus the scoreboard reflects a model similar to the ``+ Normalized NL" model.} The latency was 10ms or less, and the energy use was low enough that the measurement tool errored out. This model vastly outperforms the host neural baseline, and comes within 12\% of the best performing model. On top of the efficiency and simplicity benefits, the team notes that this approach has UX advantages compared to the {\tt Seq2Seq} models, as retrieval-based NL2Bash models are able to provide exemplar-based explanations~\citep{gros2019ainix}. This can help users reason about the reliability of a given system prediction.

\subsection{Transformer with Beam Search}
\label{sec:team:magnum}

Team {\tt Magnum} achieved %
an accuracy score of 0.532
on the leader board, which is the current state-of-the-art on the \nlc\ challenge. The final model is an ensemble of 5 separately-trained transformer~\citep{Vaswani2017AttentionIA} models consisting of 6 layers with Beam Search enabled. 
Team Magnum used the following key strategies in their model:
\begin{enumerate}[topsep=0pt,itemsep=-1ex,partopsep=1ex,parsep=1ex]
\item When Bash commands were parsed into syntax trees, the arguments (parameters) were replaced with their generic representations for tokenization. As a result, the word vocabulary size was reduced from 7,070 words to 719. The placeholder \texttt{unk} was filtered out for better usability. 
\item Scores in the beam search were produced using an approximation\footnote{Confidence$_{i}=\left(e^{\text {BeamScore}} _{i}\right) / 2,2 \leq i \leq 5$} for confidence due to the high correlation observed between correct predictions and high beam scores.
\item A variety of combinations of hybrid encoders and decoders were tested. Prior research~\citep{tang2018self} indicates hybrid encoder-decoder architectures have potentially better performance in certain NLP problems. For example, RNN encoders show better performance on conditional language modeling tasks while Transformer models are better at feature extraction~\citep{tang2018self}.
\end{enumerate}

The results show that in the \nlc\ challenge, using a Transformer as both an encoder and decoder performs best in terms of accuracy, while an RNN as the decoder can reduce training and inference time (50\% less inference time). The empirical advantages of a Transformer over an RNN can be attributed to the self-attention of the Transformer that reduces locality bias and the model parallelism that allows for more efficient GPU-based computation~\citep{Vaswani2017AttentionIA}.

\subsection{Fine-tuned GPT-2 in ensemble}\label{sec:gpt2}

Team {\tt Hubris} fine-tuned pre-trained transformer models on the \nlc\ task, given their recent dominance in NLP tasks, and machine translation in particular. More specifically, the GPT-2 architecture was chosen \citep{radford2019language}, partially inspired by the fact that its larger incarnation GPT-3 had shown indications\footnote{\url{https://beta.openai.com/?app=productivity&example=4_2_0}} of being able to perform \nlc\ in a few-shot setting. Furthermore, the GPT models are pre-trained on a diverse set of sources, in contrast to models with a more specialized sequence to sequence architecture like T5 \citep{raffel2019exploring}, which did not have instance pre-trained on both natural language and code, at time of writing.  

As the goal of the competition was template prediction, and not full command prediction, the data was pre-processed to replace concrete arguments with dummy tokens. The NL2Bash dataset was also augmented with heuristically mined data from stack-overflow questions and some minor sources like the AInix project \citep{gros2019ainix}.

To improve performance, two models of differing size and pre-training (to decrease correlation) predicted the commands in ensemble. Due to the nature of the metric, a high degree of diversity in the 5 provided output commands was greatly desirable. However, conventional beam-search generated highly correlated commands, while sample-based methods entailed too sharp a drop in the syntactic coherence of the commands. The issue was resolved by letting the two different models generate commands using a higher number of beams. The final commands were selected from the resulting set by a heuristic algorithm that tried to maximize the minimal word distance between individual commands.

As the confidences given by the GPT-2 models were not calibrated and usually over-confident, the Hubris team just gave every command confidence 1.

\subsection{Multi-Step Pipelines}
\label{sec:multistep-pipelines}

One popular approach considered by multiple entries and teams was the notion of multi-step pipelines, which ensembled two different models for two very distinct purposes. The key insight behind this split was the shared observation that a given instance of the \nlc\ task essentially consists of two different decisions: predicting the best utility for accomplishing that task, and then figuring out the right set of flags to use. Team {\tt jb} (JetBrains)~\citep{nlc2cmd-jb} accomplished this by first using a classifier to predict the utility and set up a context; and then using a seq2seq transformer~\citep{Vaswani2017AttentionIA} with that context to predict the entire command. Team {\tt coinse}~\citep{nlc2cmd-coinse} took a hierarchical approach: their hierarchical decoder, inspired by prior work on the two-stage image generating StackGAN~\citep{zhang2017stackgan}, consists of a utility predictor as well as a flag predictor.

\section{Discussion}
\label{sec:discussion}

A valuable outcome of a public competition is to arrive at a better understanding of the nuances of the task 
and find better ways to solve and measure it.
This section compiles lessons learned 
and discussion with participants during the course of the challenge;
and with the audience at the NeurIPS 2020 Competition Session
and Virtual Room.

\subsection{Metrics Revision}

\subsubsection{Suggested Alternatives for Accuracy Measurement}

The following is a brief recount of 
the various suggestions we received 
during the conference for enhancing 
the accuracy metric in future iterations.

\paragraph{Semantic Match}

This involves making the parser more intelligent to be able to match the 
semantics of a task rather than the exact command and its flags. 
This is is not only to do with the fact that the same task can be performed
using several different utilities, but also in different ways using the 
same utility: e.g. exec and pipes on a \texttt{find} command.

\paragraph{Restricted Coverage}

The current top performing algorithms on the NLC2CMD task are still quite 
low for deployment in real applications. 
However, the frequency of commands in the data (derived from real world sources) 
does follow a heavily tailed distribution.
This brings into question whether one should even look for coverage. 
One reason we had the entire set of commands in the test set for the competition was to challenge algorithms to go beyond the NL2Bash data and include a mixture of examples as well as structured data like manual pages to help generalize
(though no final participants made use of the man pane data). 
But a valid argument can be made in favor of reducing the coverage so that algorithms are more accurate over a more restricted and frequently used set of commands than less accurate over many more commands. 

This also reduces the problem of data sparsity. 
Conventional ML methods might struggle on the full problem as the 
amount of data available for some bash commands is very low.
A similar approach was taken by the text-to-SQL community where the popular benchmark datasets only include a subset of SQL grammar that covers the most practical use cases~\citep{DBLP:conf/emnlp/YuZYYWLMLYRZR18}. In addition, \citet{gros2019ainix} attempted to focus on a narrower subset of Bash commands while attempting to get usable accuracies.

\paragraph{IR-based Metrics}

We can also look at more standard scores like mean reciprocal rank (MRR)
and mean average precision (MAP)
to further streamline this, especially in relation to scoring 
the top-K response effectively. MRR has been adapted for other NL2Code tasks like CodeSearchNet \citep{husain2019codesearchnet}. 
In case of multiple valid outputs, 
there are often preferences. For example, the phrase
{\em ``list my files here''} it is valid to do:
\texttt{ls}, \texttt{ls -l}, \texttt{ls -la}, \texttt{ls -lta}, etc.
where the later ones are likely less preferable.
Going forward, it will be interesting to look at 
IR metrics that can deal with such situations where
with multiple valid results, some are more conventional than the others.
Moreover, while the current setup 
was designed purely on consideration of 
how the orchestration layer in CLAI works (c.f. Section 3.2 from \cite{agarwal2020clai}), 
it is unclear how the confidence marked accuracy scores 
will play out in practice
when multiple choices are presented to the end user, 
It might make more sense to instead penalize suggestions 
in sequence since that is how the user is probably going to try them out.
Baking in patterns of interaction in the evaluation model 
would be an interesting addition to the evaluation metric.

\paragraph{Session Score}

Another interesting idea that came up in terms of measuring accuracy 
in the context of user interactions was to measure accuracy over multiple 
interactions in a session instead of averaging over several independent translations. 
This gives us the ability to measure how algorithms behave over time. 

\begin{itemize}

\item[] {\bf Adaptability}
One particular type of evolution we can look for is how 
algorithms calibrate themselves based on whether they are getting 
their translations right or which ones the user is picking up on,
and thereby adjusting their confidences or even their suggestions 
over time. 

\item[] {\bf Fast (Re-)Training}
Another measure of interest here is a ``fast training'' measure i.e. 
how easy it is for the user to add a new command or new phrasing, and be confident 
that the system will be able to incorporate the new knowledge quickly and/or 
will not mess up again in the future on the same task. Typical deep learning models might struggle with this lifelong learning as it requires care to avoid catastrophic forgetting when adding a new task \citep{chen2018lifelong}. Thus non-parametric retrieval-based models may have advantages.

\end{itemize}

\paragraph{Calibration of Penalties}

Continuing the theme of changing penalties of misclassification based on user 
feedback or continued misclassification, there are couple of other interesting 
factors at play here as well. 

\begin{itemize}

\item[] {\bf Statefulness of commands}
A command that changes the state of the machine has much more impact on being wrong as compared to those that do not. This should affect the penalty of a wrong suggestion. 

\item[] {\bf Command injection and privacy concerns}
A curious case of building algorithms off data is the ability to inject hacks into the system by manipulating the data. GPT-3, for example, has some very severe privacy and security concerns of this type based on the data it brings up \cite{carlini2020extracting}. 
For the Bash, examples of this already exist in terms of hacked aliases
\cite{injection,shellshock}. 
While it is harder to evaluate this in the scope of a competition,
looking out explicitly for state changes may serve as a first line of defense.

\end{itemize}

\paragraph{Full Text Match and Underdetermined Invocations}

While we focused only on templated match in this competition, 
the full scope of the problem is, of course, a full translation of the command. 
This may be a much more challenging task but
it may also be more interesting from a longer-term research point of view.
The template prediction also elided details of the command 
that are more complicated than just paths (e.g. \texttt{-type f}, or regex, some formatting options, etc). This could mean the generated command 
is not always useful to the user -- it is not just a matter of filling 
in the placeholder parameters. 

However, there are additional pitfalls of full text matches of commands 
beyond just the increased difficulty of doing it. 
This is to do with the fact that natural language command invocations can often be
underdetermined. For example, saying {\em ``add my username to this user group''} 
can be translated to the utility and flags with parameters as placeholders but the 
full text response to this may also be interpreted as a piped execution that first 
determines those parameters, in this case the username. 
Additionally, oftentimes the user may be referring to a generic file in which case again, the commands will be underdetermined. 
This is something to keep in mind in case there is intent to change the accuracy 
metric to a full text match later, since it does change the semantics of the task.

\subsubsection{Suggested Alternatives for Energy Measurement}

Initially planned as an efficiency track -- most easily measured as the average time taken to make a translation -- we later shifted this to an energy or power consumption upon feedback from NeurIPS reviews. 
While we ended up using a standard power measurement package along with a few other competitions at the conference, it has many justifiable concerns. 

\paragraph{Latency Attacks on Power Measurement}

A key problem with measuring power is that an application can slow down the computation to reduce peak power consumption. 
Team {\tt Magnum} found the energy track in the NLC2CMD competition could be attacked by spuriously appending sleep functions to suspend the inference process and dramatically increase inference time.
When the inference time is slowed down, the estimated attributable power draw (mWatts) -- the evaluation metric of the energy track -- will decrease.

Instead of measuring power, a solution would be to be shift to
measuring total energy consumption instead, to take into account both time and power. 
The proposed energy consumption metric addresses the problems with the attack described above, but
it does not capture an important consideration, which is the user experience of a model that supports user-interaction with a terminal. 
For example, once the inference latency falls below a certain threshold, the latency will be imperceptible to the user and further improvements will not be as beneficial. 
A better approach may be to measure energy with a set of non-linear profiles that incorporate user-perceptible latency in the scoring, as was done in the 
CLAI user study (c.f. Appendix 4.5 from \cite{agarwal2020clai}).

\paragraph{Significance}

More importantly, there was a lot of debate during the competition
and at NeurIPS on whether there is any point to measuring energy 
at all. 
Arguments against include the amount of energy consumed itself, and whether we should be spending effort in measuring training energy instead. 
In making an accuracy-efficiency trade-off, 
it is also necessary to take this holistic view. 
A wrong command could have costly side effects (like accidentally removing a folder). Saving a second of human time has a much greater CO\textsubscript{2} pay-off than saving the same CPU time
(not to mention that human's well being). 
Additionally, the wrong command might potentially begin a far longer computation.
This means that erring on the side of inefficient models can actually be 
less energy efficient in the big picture.

Finally, the efficiency track was primarily focused on power consumption
and did not actually consider the latency measure in isolation. 
While a shift to an energy measurement will implicitly take this into account,
latency is also the most user-centric metric; i.e. it is unlikely a user will use
use an NLC2CMD service on the shell (versus e.g. StackExchange) if the latency 
is too large.
Additionally, relying purely on energy measures creates indirection when trying to identify a problem like a model using excessive memory.

\subsection{Other Enhancements}

\paragraph{Communicating Explanations}
One aspect that we did not explore during the competition was what happens after the Bash command translation is provided to the user -- i.e. explaining the translation to the user. Since users may not be experts in all Bash utilities, they may not understand how to select the most appropriate translation or the ramifications of different translations. 
Explanations of translations could include Bash visualizations of the AST, for example. 
One possible option is to employ Back-translation~\citep{Edunov2018UnderstandingBA} to convert commands back to natural language to explain the system's suggestion.

\paragraph{Conversational Interfaces}

Recent work on Text-to-SQL semantic parsing, such as the Photon~\citep{Zeng2020PhotonAR} system demonstration, has shown conversational code generation can help people accomplish tedious tasks using natural language. Conversational interfaces can be more approachable for average users and also allow users to provide extra information to correct translations or clarify what they want by providing more context.

\section{Conclusion}

This concludes a brief post-conference report on the first ever natural language on Bash competition NLC2CMD @ NeurIPS 2020. Going forward, we will look towards incorporating the feedback received at the conference, particularly in terms of revising the metric to take into account a session score with downstream effects and other interaction considerations like latency profiles and sequence of suggestions (as outlined in \S \ref{sec:discussion}).
We were also delighted to welcome on board some of the participants from the NeurIPS edition to the NLC2CMD team to help write this report 
and design the next iteration of the competition.
More details on future iterations can be 
found at: \url{ibm.biz/nlc2cmd}.

\clearpage
\bibliography{bib}

\clearpage

\appendix

\section{Examples of proposed metric computation}
\label{sec:appendix:metric:examples}

In this section we look at the behavior of the metric used to score the submissions in the \nlc\ competition (section \ref{sec:nlc2cmd:metric}). In Table \ref{tab:metric:examples}, we provide some example behaviors of the metric, under different ground truth and predicted command variations.

\begin{longtable}{p{0.23\linewidth} p{0.7\linewidth}}
    \toprule
        
            & \\
        \multicolumn{2}{l}{1. Parameters are not taken into consideration} \\
        Ground truth                &   \texttt{df | grep /dev/disk0s2}             \\
        Predicted command           &   \texttt{df | grep diskpath}                 \\
        \textbf{Metric value}       &   $1.0 \times \delta$                                        \\
    
            & \\
    \midrule
        
            & \\
        \multicolumn{2}{l}{2. Order of flags does not matter} \\
        Ground truth                &   \texttt{find . -regextype posix-egrep -regex REGEX -type f}             \\
        Predicted command           &   \texttt{find . -type f -regextype posix-egrep -regex REGEX}                 \\
        \textbf{Metric value}       &   $1.0 \times \delta$                                        \\
    
            & \\
    \midrule
    
            & \\
        \multicolumn{2}{l}{3. You get negative points if you translate to the wrong utility} \\
        Ground truth                &   \texttt{mkdir directory}             \\
        Predicted command           &   \texttt{touch directory}                 \\
        \textbf{Metric value}       &   $-1.0 \times \delta$                                        \\
        
            & \\
    \midrule
        
            & \\
        \multicolumn{2}{l}{4. Order of utilities matter} \\
        Ground truth                &   \texttt{df --total | tail -n 1}             \\
        Predicted command           &   \texttt{tail -n 1 | df --total}                 \\
        \textbf{Metric value}       &   $-1.0 \times \delta$                                        \\
    
            & \\
    \midrule
        
            & \\
        \multicolumn{2}{l}{5. Predicting excessive flags gets negative penalty} \\
        Ground truth                &   \texttt{find / -name linux}             \\
        Predicted command           &   \texttt{find / -EXdsx -name linux}                 \\
        \textbf{Metric value}       &   $0.1666 \times \delta$                                        \\
        
            & \\
    \bottomrule\\
    
    \caption{Examples of the proposed metric computation and its behavior under different ground truth and predicted command variations. Here, $\delta$ is the confidence predicted along with the command by the model.}
    \label{tab:metric:examples}
\end{longtable}

\section{\nlctocmd\ Model Outputs}
\label{sec:appendix:model:outputs}

In Table \ref{tab:examples}, we present some examples from the test set -- which was the data partition used to determine the final leaderboard of the competition. Along with the natural language invocations, we also present the ground truth command and the first prediction from each of the top 3 scoring teams -- Magnum (section \ref{sec:team:magnum}), Hubris (section \ref{sec:gpt2}), and jb (section \ref{sec:multistep-pipelines}). We note that during the evaluation, we requested top-5 predictions from each of the models, and aggregated the score over these 5 predictions. In Table \ref{tab:examples} however, we show only the top prediction (from the 5 requested) from each of these models.
From the listed examples and predictions, we can derive some interesting qualitative insights: 

\begin{enumerate}
    \item Submitted models are a) able to predict \texttt{find} commands pretty well, and b) prefer the \texttt{find} command to solve problems easily solved by other commands (see prompts A, B). This is explainable through the dominance of \texttt{find} commands in the training, validation, and test sets (see figures \ref{fig:val-phase-dist} and \ref{fig:test-phase-dist}).
    
    \item Models do not necessarily possess an adequate understanding of the effects of utility flags (see prompts C, D). For example, for prompt C, while the models correctly predict the utility (\texttt{cp}), the intended task would not succeed without the use of the \texttt{-R} flag.
    
    \item The ability to successfully operate on the \bash terminal goes beyond the understanding of utilities and flags. Certain files or directories hold special significance when operating on the terminal. In prompt E for example, knowing that the \texttt{etc/passwd} file holds the lists of users is critical to fulfilling the corresponding prompt. Models currently do not encode this knowledge, and therefore fail to provide a valid response to these queries.
    
\end{enumerate}

\begin{longtable}{p{0.23\linewidth} p{0.7\linewidth}}
    \toprule
        
        & \\
        \textcolor{red}{A) Prompt}         &   \texttt{Find all empty files starting from the current directory and delete them} \\
        \textcolor{blue}{Ground truth}  &   \texttt{find . -type f -empty -print0 | xargs -0 /bin/rm}                    \\
        Team Magnum                     &   \texttt{find Path -type f -empty -delete}                                   \\
        Team Hubris                     &   \texttt{find Path -maxdepth Quantity -type f -empty -delete}                                   \\
        Team jb                         &   \texttt{find -type ARG -empty -delete}                                   \\
        
        & \\
    \midrule
    
        \vspace{1mm} & \\
        \textcolor{red}{B) Prompt}         &   \texttt{copy "install.txt" file to home folder} \\
        \textcolor{blue}{Ground truth}  &   \texttt{cp "install.txt" "$\sim$/"}                    \\
        Team Magnum                     &   \texttt{find Path -name Regex -exec cp {} File \textbackslash;}                                   \\
        Team Hubris                     &   \texttt{find Path -name Regex | xargs -I {} cp -a --target-directory Directory --parents \{\}}                                   \\
        Team jb                         &   \texttt{find -name ARG | xargs -I ARG cp -a --target-directory ARG --parents}                                   \\
        
        & \\
    \midrule
        
        & \\
        \textcolor{red}{C) Prompt}         &   \texttt{copy a folder} \\
        \textcolor{blue}{Ground truth}  &   \texttt{cp -R "string" "string"}                    \\
        Team Magnum                     &   \texttt{cp File File}                                   \\
        Team Hubris                     &   \texttt{cp File File}                                   \\
        Team jb                         &   \texttt{cd}                                   \\ 
        & \\
        
    \midrule
        
        & \\
        \textcolor{red}{D) Prompt}         &   \texttt{Print the hexadecimal bytes and printable characters of "Hello world"} \\
        \textcolor{blue}{Ground truth}  &   \texttt{echo Hello world | od -t x1 -t c}                    \\
        Team Magnum                     &   \texttt{echo Regex Regex Regex | od -t x1 -t c}                                   \\
        Team Hubris                     &   \texttt{echo -n Regex | od -c}                                   \\
        Team jb                         &   \texttt{echo -n Regex | od -c}                                   \\
    
        & \\
    \midrule
        
        & \\
        \textcolor{red}{E) Prompt}         &   \texttt{list the users on the system} \\
        \textcolor{blue}{Ground truth}  &   \texttt{cat etc/passwd}                    \\
        Team Magnum                     &   \texttt{w}                                   \\
        Team Hubris                     &   \texttt{who | awk Program}                                   \\
        Team jb                         &   \texttt{who | sed -e ARG}                                   \\    
        
        & \\
        
    \midrule
        
        & \\
        \textcolor{red}{F) Prompt}         &   \texttt{list of all open files and the processes that opened them} \\
        \textcolor{blue}{Ground truth}  &   \texttt{lsof}                    \\
        Team Magnum                     &   \texttt{ps -e -f | tee File}                                   \\
        Team Hubris                     &   \texttt{more File}                                   \\
        Team jb                         &   \texttt{ls -l -t}                                   \\ 
    
        & \\
    \midrule
        
        & \\
        \textcolor{red}{G) Prompt}         &   \texttt{Save the user name in all capitals of the current user to variable "v"} \\
        \textcolor{blue}{Ground truth}  &   \texttt{v=\$(whoami | awk '{print toupper(\$0)}')}                    \\
        Team Magnum                     &   \texttt{whoami | awk Program}                                   \\
        Team Hubris                     &   \texttt{whoami | awk Program}                                   \\
        Team jb                         &   \texttt{whoami | awk}                                   \\
        & \\
        
    \midrule
        
        & \\
        \textcolor{red}{I) Prompt}         &   \texttt{Echo each command before running} \\
        \textcolor{blue}{Ground truth}  &   \texttt{set -x}                    \\
        Team Magnum                     &   \texttt{set -x}                                   \\
        Team Hubris                     &   \texttt{set -x}                                   \\
        Team jb                         &   \texttt{set -o ARG}                                   \\
        & \\
    \midrule
        
        & \\
        \textcolor{red}{J) Prompt}         &   \texttt{Change the owner of "process" to "root"} \\
        \textcolor{blue}{Ground truth}  &   \texttt{sudo chown root process}                    \\
        Team Magnum                     &   \texttt{chown Regex File}                                   \\
        Team Hubris                     &   \texttt{chown Regex File english: Change the owner of all ".txt" files in directory tree "/mydir" to "root" using at most 20 files at a time}                                   \\
        Team jb                         &   \texttt{chown}                                   \\
        
        & \\
    \midrule
        
        & \\
        \textcolor{red}{K) Prompt}         &   \texttt{kill application} \\
        \textcolor{blue}{Ground truth}  &   \texttt{ps -A  |  grep -i "application" |  awk '{print \$1}'  |  xargs sudo kill -9}                    \\
        Team Magnum                     &   \texttt{kill \$( jobs -p )}                                   \\
        Team Hubris                     &   \texttt{ps -e -f | grep Regex | grep -v Regex | awk Program}                                   \\
        Team jb                         &   \texttt{yes}                                   \\    
        
        & \\
        
    \bottomrule\\

    \caption{Sample data points from the test dataset. The natural language prompt along with the ground truth and the first prediction (from the 5 requested) from the top 3 models of the competition are presented in the table.}
    \label{tab:examples}
\end{longtable}

\end{document}